\documentclass[]{spie}
\usepackage{graphicx,color}
\usepackage{tabularx}
\usepackage{multirow}
\usepackage{epstopdf}
\usepackage{algorithmic}
\usepackage{algorithm}
\usepackage{amsmath,amssymb,amstext}
\usepackage{subfig}
\usepackage{url}
\usepackage{colortbl}
\usepackage{fancyheadings}
\newcommand{\CL}{\cellcolor[gray]{0.8}}

\title{Anatomy-Aware Measurement \\ of Segmentation Accuracy}

\author[a]{H.R.Tizhoosh}
\author[b]{A.A. Othman}
\affil[a]{KIMIA Lab, University of Waterloo, 200 University Avenue West, Waterloo, Canada}
\affil[b]{Dept. of Information Systems, Computers and Informatics, Suez Canal University, Egypt}

\authorinfo{E-mails: tizhoosh@uwaterloo.ca, a.othman@ci.suez.edu.eg}

\begin{document}
\maketitle

\begin{abstract}
Quantifying the accuracy of segmentation and manual delineation of organs, tissue types and tumors in medical images is a necessary measurement that suffers from multiple problems. One major shortcoming of all accuracy measures is that they neglect the anatomical significance or relevance of different zones within a given segment. Hence, existing accuracy metrics measure the overlap of a given segment with a ground-truth without any anatomical discrimination inside the segment. For instance, if we understand the rectal wall or urethral sphincter as anatomical zones, then current accuracy measures ignore their significance when they are applied to assess the quality of the prostate gland segments. In this paper, we propose an anatomy-aware measurement scheme for segmentation accuracy of medical images. The idea is to create a ``master gold'' based on a consensus shape containing not just the outline of the segment but also the outlines of the internal zones if existent or relevant. To apply this new approach to accuracy measurement, we introduce the anatomy-aware extensions of both Dice coefficient and Jaccard index and investigate their effect using 500 synthetic prostate ultrasound images with 20 different segments for each image. We show that through anatomy-sensitive calculation of segmentation accuracy, namely by considering relevant anatomical zones, not only the measurement of individual users can change but also the ranking of users' segmentation skills may require reordering.
\end{abstract}

\section{Description of Purpose}
Firefighters battling to extinguish a burning city block manage to put out the flames in 95\% of the empty buildings. Many residents, however, die in the remaining 5\% of the buildings. 

What would we feel about the performance of those firefighters if this horrible scenario were real news? Does the number ``95\%'' really mean anything? Wouldn't we have preferred to let the 95\% of empty buildings simply burn down, and instead, focus on those 5\% with people living in them?  This firefighting metaphor should illustrate the magnitude of the problem when we deal with the measurement of accuracy of organ, tumor and tissue segments in medical applications. Generally, we do focus on the whole segment without paying attention to any anatomically or pathologically significant zones inside the segment. Accuracy and its measurement is a very challenging topic in medical image analysis. Often, one can speak of accuracy when there exists a reference line, a benchmark instance, against which the current estimate or guess can be compared. We usually call this reference either ``ground-truth'' or, sometimes rather loosely, ``gold standard'' images. Ground-truth images are manual delineations created by the medical expert (e.g., radiologist, oncologists). The results of any segmentation algorithm, automated or not, can then be quantified via comparison with this ground-truth image. The accuracy of manual delineations can be measured against consensus segments among multiple experts (gold standard image). Hence, algorithms are accurate if their segments do overlap with what experts expect. That is the case in all validation procedures when we test the performance of software algorithms or the quality of manual delineations. In other words, we treat all pixels of a segment in the same way although, in many clinical cases, there are clearly different zones that are of lower or higher significance for the task at hand. As an example, when we are segmenting prostate glands for radiation treatment, the \emph{rectal wall} is a critical zone for which the segment should exhibit highest accuracy possible. Another example is when we examine breast ultrasound lesions for diagnostic purposes. Here when the mass is mostly segmented correctly but some ``spiculations'' are missed, this can completely change the lesion classification based on BI-RADS guidelines.

Our idea is to establish a zone-sensitive, or anatomy-aware accuracy measurement that can take into account anatomical or pathological a-priori knowledge and incorporate it into the accuracy measurement.   

\section{The Methods}
There is a vast literature on evaluation of segmentation results \cite{Popovic2007,Shepherd2012,Zou2004,Correia2000,Chang2009,Udupa2006}. The problem of validating the segmentation accuracy in medical image analysis is apparently that we look at the entire segment without any internal discrimination, meaning that some important zones inside the segment are completely ignored. What is the solution? It seems that we cannot develop any solution unless those ``significant zones'' inside the segment are defined prior to the calculation. But that means we have to ask the medical expert to highlight the zones in every segment, and this can be a very tedious task and hence an infeasible requirement. Keeping in mind that ground-truth segments by at least one expert must be available for any type of accuracy measurement, we cannot put additional burden of delineating the zones in individual ground-truths on the expert. So what is the solution?

\begin{figure}[htb]
\begin{center}
\includegraphics[width=0.5\textwidth]{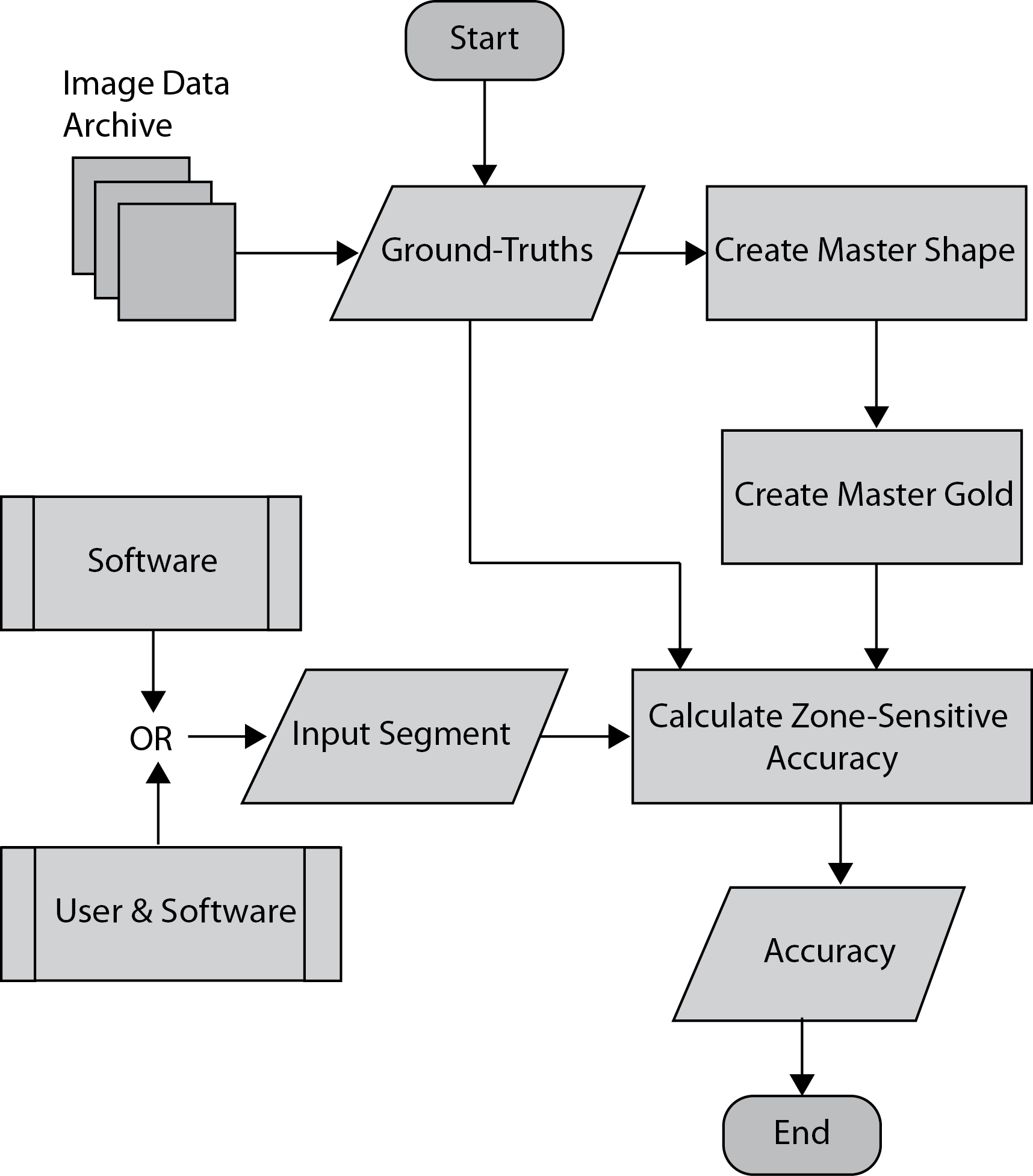}   \\
\caption{In addition to segments and ground-truths, a master gold should be created to calculate the zone-sensitive accuracies. The master gold depicts a generic shape, called master shape, with defined zones. The segment can come from experts and/or software. Ground-truth images come from one or multiple experts.}
\label{fig:FlowChart}
\end{center}
\end{figure}

The zones have to be highlighted in a ``master shape'', a general or statistical shape that represents the expected shape appearance of the organ or tumor. Of course, such an approach can only address the cases with more or less regular shapes, e.g., organs and compact masses such as cysts and nodules. As well, it would need to be done only once in order to not create additional work for the clinical experts.  A master shape with zones inside would then constitute a ``master gold''. Every time that we have a segment and corresponding ground-truth, we can map the zones from the master gold to the current ground-truth and subsequently to the segment. This finally enables us to perform zone-sensitive accuracy measurements provided we also have some zone-sensitive accuracy measures (if we extend existing ones to become aware of zonal anatomy within the segment) to capture the compound accuracy. The outline of this idea is illustrated in Figure \ref{fig:FlowChart}. 
    
Measuring accuracy of segmentation is generally possible if a ground-truth segment is available. This is most of the time a manual segmentation by an expert, against which the accuracy of any segments can be measured. If there are several manual segmentations by multiple experts available for the same image, then one may build a \emph{consensus contour} to serve as gold standard image. 

\begin{figure}[htb]
\begin{center}
\includegraphics[width=0.4\textwidth]{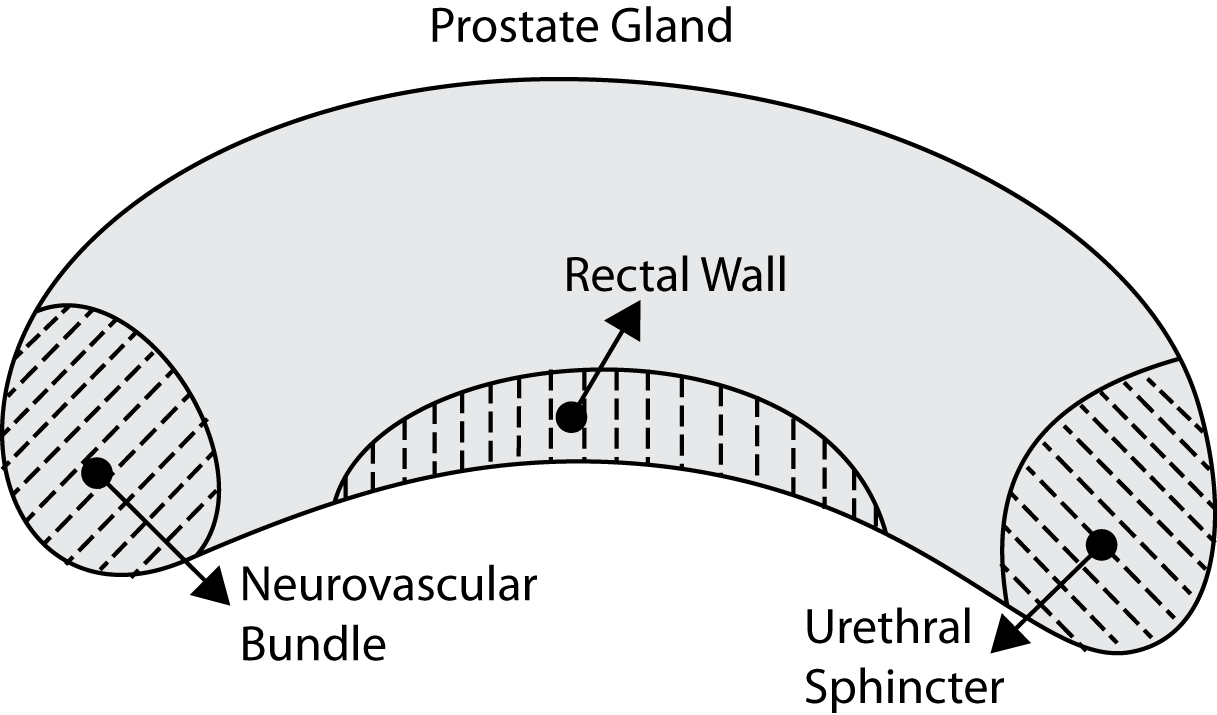}   \qquad
\includegraphics[width=0.4\textwidth]{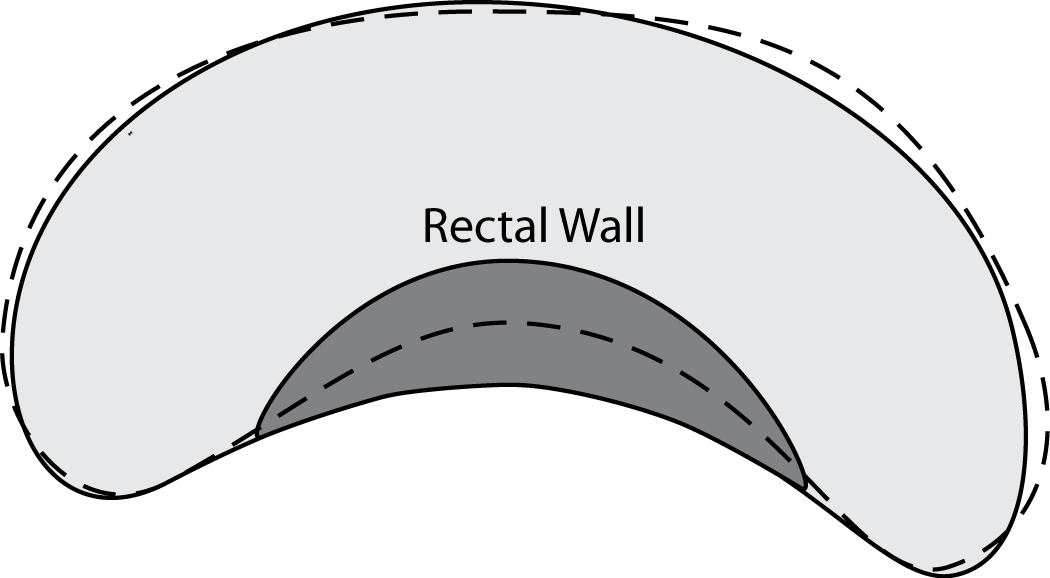}   \vspace{0.07in}
\caption{Significant zones within the prostate gland are generally ignored by existing accuracy measures (left). Hence, segments (dashed outline) may receive high accuracy values even though they miss a considerable potion of the rectal wall (right, dark gray).}
\label{fig:ProstateZones}
\end{center}
\end{figure}

Given the segment $S$ and the ground-truth $G$, the Jaccard index $J(S,G)$, sometimes called the area overlap and occasionally called Tanimoto index, can be calculated as follows \cite{Jaccard1912}: 
\begin{equation}
J(S,G) = \frac{|S\cap G|}{|S\cup G|}.
\end{equation}
Given the segment $S$ and the ground-truth $G$, the Dice coefficient $D(S,G)$ can be calculated as follows \cite{Dice1945}: 
\begin{equation}
D(S,G) = \frac{2|S\cap G|}{|S|+|G|}.
\end{equation}
One can show that $J=D/(2-D)$ and $D=2J/(1+J)$, hence $J\!<\!D$. It is obvious that $S$ can come from an algorithm in which case $G$ is the ground-truth from one or multiple users. As well, $S$ can be manual delineation by an expert whereas $G$ is then gold standard as consensus among multiple experts. For instance, when segmenting the prostate gland, one has to actually pay more attention to some specific zones such as the rectal wall, neurovascular bundle and urethral sphincter (Figure \ref{fig:ProstateZones}, left). In many cases, a segment may have a large overlap with the ground-truth but may not be accurate enough in significant zones (Fig. \ref{fig:ProstateZones}, right). The accuracy of such segments should be penalized according to the zonal accuracy.

Any anatomy-aware accuracy measure $A^*$ with higher emphasis on zonal accuracy should hence be the extended version of an existing accuracy measure $A$ (for instance, Jaccard or Dice) when the zonal accuracy $A_Z$ is given and a convex combination can be built with    
\begin{equation}
A^* = \alpha A + (1-\alpha) \times A_Z,
\end{equation}
\noindent where $\alpha \in [0,1]$. Of course, if there are $N_Z$ zones, then a representative zonal accuracy among the accuracies $A_{Z_1},A_{Z_2},\dots,A_{Z_{N_Z}}$ should be calculated. One may, conservatively chose 
\begin{equation}
A^* = \alpha A + (1-\alpha) \times \underset{i}{\min}~ A_{Z_i}.
\end{equation}
However, the trade-off value $\alpha$ that determines how significant the zones are relative to the overall segment would pose another adjustment problem which is clearly not desirable. But to further investigate the establishment of a new accuracy measure, let us look at the extreme values for such trade-off parameter. In case $\alpha \rightarrow 1$  the influence of the zonal accuracy, expectedly, disappears. For $\alpha \rightarrow 0$ the zonal accuracies become dominant. However, this indicates a problem that in case the accuracy of overall segment is not high enough it would not be meaningful to pay attention to zonal accuracies. Considering these thoughts, we can establish
\begin{equation}
A^* = A^2 + (1-A) \times \underset{i}{\min}~ A_{Z_i} \quad \textrm{if}\quad A\geq A_{\min},
\label{eq:generalacc}
\end{equation}
\noindent where $A_{\min}$ is the minimum required segment accuracy for the application at hand. For instance, if the expert/software sets $A_{\min} = 75\%$ that means the zonal accuracies will only be considered via $A^*$ if the overall segment accuracy is at least $75\%$.  Depending on the critical role of segmentation, any segment with $A\!<\!A_{\min}$ may be rejected as unacceptable.

Hence, to make Dice coefficient anatomy-aware, one may use 
\begin{equation}
D^{*}_{1} = D^2+(1-D)\times \underset{i}{\min}~ D_{Z_i}.
\end{equation}
Or alternatively, one may modify the core definition of the Dice coefficient to incorporate zones (TP=true positive, FP=false positive, FN=false negative):

\begin{equation}
D^{*}_{2} = \frac{2(\sum_i^n \textrm{TP}+\sum_i^{N_Z} \textrm{TP}_{Z_i})}{\sum_i^n (2\textrm{TP}+\textrm{FP}+\textrm{FN}) + \sum_i^{N_Z}  (2\textrm{TP}_{Z_i}+\textrm{FP}_{Z_i}+\textrm{FN}_{Z_i})}.
\end{equation}

Analogously, the Jaccard index can be extended as follows:
\begin{equation}
J^{*}_{1} = J^2+(1-J)\times \underset{i}{\min}~ J_{Z_i}.
\end{equation}

The Jaccard extension can also occur by changing the core definition: 
\begin{equation}
J^{*}_{2} = \frac{\sum_i^n \textrm{TP}+\sum_i^{N_Z} \textrm{TP}_{Z_i}}{\sum_i^n (\textrm{TP}+\textrm{FP}+\textrm{FN}) + \sum_i^{N_Z} (\textrm{TP}_{Z_i}+\textrm{FP}_{Z_i}+\textrm{FN}_{Z_i})}.
\end{equation}

\textbf{Extracting the Master Shape (Algorithm \ref{alg:GeneralShape}) --} In order to calculate the extended accuracy measures, one apparently needs a very different approach to segmentation evaluation. Using existing ground-truth images $G_i$, we calculate a general (master) shape $M_S$. In addition to a desired minimum accuracy $A_{\min}$, the expert  has to determine the number of zones $N_Z$. As well, the threshold $T_{\textrm{shape}}$ needs to be set which determines the consensus level for thresholding the accumulated ground-truths (line 3, Algorithm \ref{alg:GeneralShape}) (all pixels with at least $T_{\textrm{shape}}$ overlap among segments will belong to the consensus segment).  One may use algorithms like STAPLE \cite{Warfield2004}, however this failed in working with a large number of segments in our experiments such that we we were forced to use our simple method to extract the master shape $M_S$.

\begin{algorithm}[htb]
\caption{Extract the General Segment Shape $M_S$}
\label{alg:GeneralShape}
\begin{algorithmic}[1]
\STATE  User sets the shape threshold $T_{\textrm{shape}}$ (e.g., $T_{\textrm{shape}}=50\%,60\%,\dots$).
\STATE  Load the available gold images $G_1,G_2,\dots, G_n$.
\STATE  Create cumulative image: $C_G \leftarrow \sum_{i=1}^n G_i$.
\STATE  Get the master shape: $M_S \leftarrow$ Binarize $C_G$ with threshold $=  (n \times \frac{T_{\textrm{shape}}}{100})$
\STATE  Save $M_S$.
\end{algorithmic}
\end{algorithm}

\textbf{Creating the Master Gold (Algorithm \ref{alg:MasterGold}) --} In a second phase, one would need to let the expert delineate $N_Z$ zones in the master shape $M_S$ using $N_P$ points (clicks) per zone to create the master gold $M_G$. We implemented  Algorithm \ref{alg:MasterGold} to perform this phase, however, the zones can be delineated using any available image editor. Also one has to bear in mind that the creation of the master gold is a one-time task and generally does not need to be repeated.

\begin{algorithm}[htb]
\caption{Create the Master Gold $M_G$ from Master Shape $M_S$ by acquiring the zones from user}
\label{alg:MasterGold}
\begin{algorithmic}[1]
\STATE Load the master shape $M_S$.
\STATE Set the number of (clicks) points $N_P$
\FOR{i = 1 : $N_Z$} 
	\FOR{j = 1 : $N_P$}
		\STATE Ask the user to select a point $P_i = (x_j, y_j)$.
		\IF{$P_i$ is close to the $M_G$ contour}
			\STATE Adjust $P_i$ to be on the contour.
			\STATE Save $P_i$
		\ELSE
			\STATE Save $P_i$ as a middle point
		\ENDIF
	\ENDFOR
\STATE Use the $N_P$ points to create a curve $C_i$.
\STATE $Z_i \leftarrow $Fill in the $i$-th zone bounded by $C_i$ and $M_S$ border.
\STATE $M_G \leftarrow M_S + Z_i$
\STATE Save the coordinate of the zone $P_i=(x_j, y_j)$.
\ENDFOR
\STATE Save $M_G$ 
\end{algorithmic}
\end{algorithm}

A soon as a master gold $M_G$ is available, one can start calculating the accuracy of segments using the ground-truths $G$ provided the zones depicted in $M_G$ can be aligned with corresponding points in the $i$-the ground-truth $G_i$ and the segment $S_i$. Whereas the master gold $M_G$ is one image and universally available for all images, every image $I_i$ with the segment $S_i$ has, as usual, its own ground-truth $G_i$ for evaluation or training purposes. 

\begin{figure}[tb]
\begin{center}
\vspace{0.05in}
\includegraphics[width=0.6\textwidth]{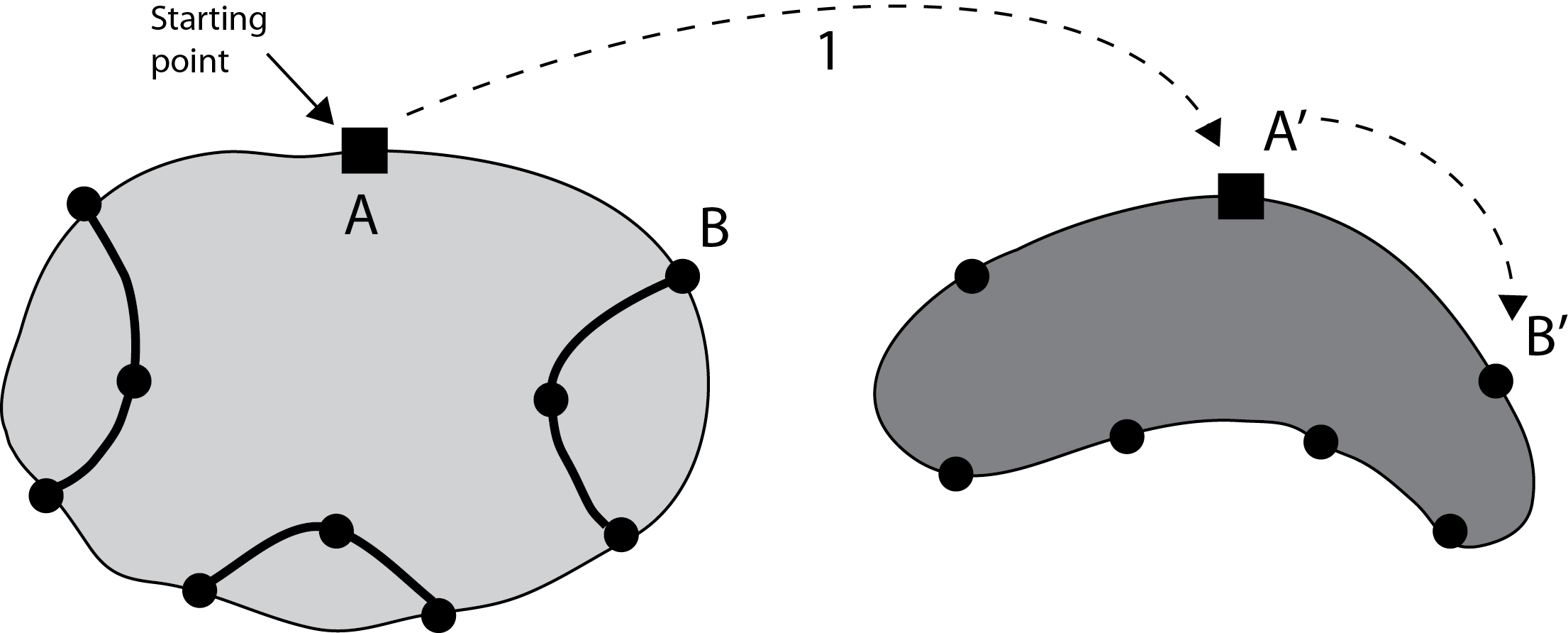}   \vspace{0.07in}
\caption{Point Mapping: The salient points of the zones defined in the master gold $M_G$ (left) are mapped into the current ground-truth and segment (right).}
\label{fig:PointMapping}
\end{center}
\end{figure}

\textbf{Mapping instead of Registration (Algorithm \ref{alg:Map2Segment}) --} Finding the correspondent pixels in $G_i$ and consequently in $S_i$, given the zonal coordinates in $M_G$, seems to be a typical ``registration'' task. However, based on our experimental results we decided to not use registration algorithms for this purpose. The non-rigid registrations we tested were both time-consuming (which may not be a critical drawback) and inaccurate. Whereas one may use a specific registration algorithm in context of a familiar segmentation task, we do provide a quasi-non-rigid mapping procedure that is very fast, due to its simplicity, and can handle small irregularities quite easily.  For this, first we do fix some points on the contour of the master gold (see Algorithm \ref{alg:MasterGoldPoints} in Appendix) and then map them to the ground-truth (see Algorithm \ref{alg:Map2Gold} in Appendix) and segment (Algorithm \ref{alg:Map2Segment}; see Figure \ref{fig:PointMapping}).

\begin{algorithm}[tb]
\caption{Map Zones to the Segment (see Algorithm \ref{alg:Map2Gold} and Figure \ref{fig:PointMapping})}
\label{alg:Map2Segment}
\begin{algorithmic}[1]
\STATE Load the current segment $S$ and the Master Gold $M_G$.
\FOR{i = 1 : $N_Z$}
	\IF{the zone at the right or the left}
		\STATE Apply the $x$-values at the $x$-axis on $P_N$ to calculate the $y$-values.
	\ELSE 
		\STATE Apply the $y$-values at the $y$-axis on $P_N$ to calculate the $x$-values.
	\ENDIF
	\STATE Draw a curve using $x$ and $y$ values.
	\STATE Fill in the area under the curve that belong to the segment to create the zone.
	\STATE Remove any part of the curve that fall out of the segment.
	\STATE Save the coordinates of the zone $(x_{S_i}, y_{S_i})$.
\ENDFOR
\end{algorithmic}
\end{algorithm}

\section{Results} 
\label{ExpRes}
Only organs and regular-shaped anomalies (cysts, nodules etc.) are considered. We further assume that there is at least one expert who has created ground-truth segments for each image and there is at least one expert who can mark anatomically meaningful zones with higher significance for segmentation. And finally we assume that the zones always touch the boundary of the segment.  

\subsection{Image Data: Synthetic TRUS Images} 
It is a challenge to validate any approach to segmentation. One has to measure the accuracy of the segment $S$ against ground-truth images. Ideally, if we have many users available to segment images, we can build ``consensus segments'', or \emph{gold standard}, to make more reliable measurements. Of course, this is usually not feasible with real images, for which there is no gold standard. Hence, we generated synthetic images whose gold segments were known a priori. For this reason, we used synthetic images that simulate transrectal ultrasound (TRUS). 

TRUS images of prostates may be used to both diagnose and treat prostate diseases such as cancer. Starting with a set of prostate shapes $P_1, P_2,\dots, P_m$, we created random segments $G_i$ through combinations of those priors, adding noise along with random translations and rotations, and we distorted the results with speckle noise and shadow patterns. Each image $I_i$ is thus created from its gold $G_i$. Consequently, we can simulate $k$ user delineations $S_i^1, S_i^2,\dots, S_i^k$ by manipulating $G_i$ via scaling, rotation, and morphological changes, and we can simulate edits by running active contours with variable user-simulating parameters. The variability of user delineations was simulated according to several factors: error probability ($[0,0.05]$), anatomical difficulty ($=0.2$ out of $[0, 1]$), and the scaling factor for morphology (form $1\!\times\!1$ to $21\!\times\!21$). The user was modelled according to the level of experience (a random number from $(0,1]$), the user's attention  (a random number from $[0,1]$), and the user's tendencies in terms of the segment size (a random number from $[-1,1]$), whether tending to draw contours that are relatively small ($\rightarrow\!-1$) or large ($\rightarrow\!+1$). 

\begin{figure}[htbp]
\begin{center}
\vspace{0.05in}
\includegraphics[width=1.2in,height=1.2in]{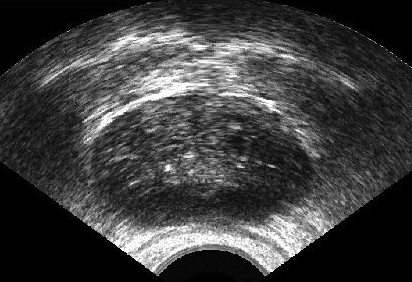}
\includegraphics[width=1.2in,height=1.2in]{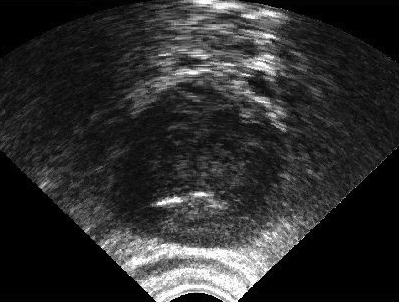}
\includegraphics[width=1.2in,height=1.2in]{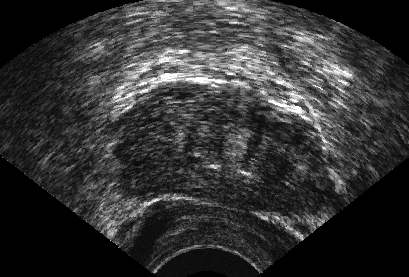} 
\includegraphics[width=1.2in,height=1.2in]{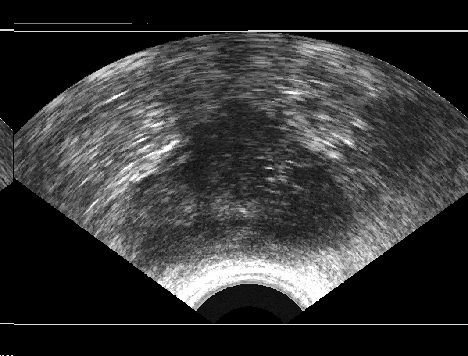} 
\includegraphics[width=1.2in,height=1.2in]{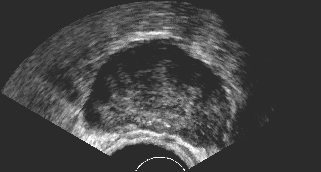} \\ \vspace{0.05in}
\includegraphics[width=1.2in,height=1.2in]{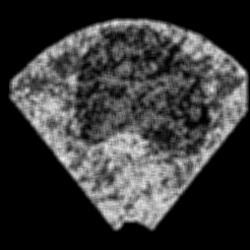}
\includegraphics[width=1.2in,height=1.2in]{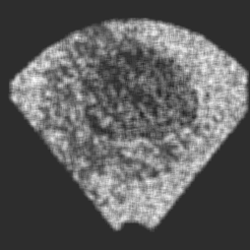}
\includegraphics[width=1.2in,height=1.2in]{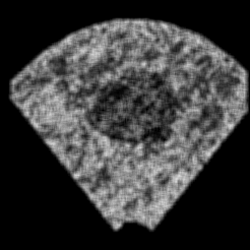} 
\includegraphics[width=1.2in,height=1.2in]{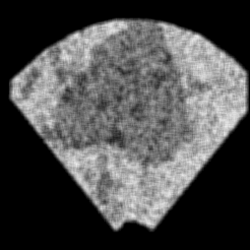} 
\includegraphics[width=1.2in,height=1.2in]{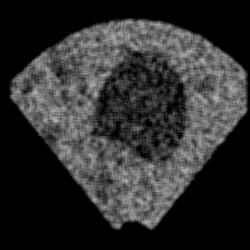} \vspace{0.07in}
\caption{Sample TRUS images (top) and simulated images (bottom).}
\label{fig:TRUSsample}
\end{center}
\end{figure}

\begin{figure*}[htbp]
\begin{center}
\includegraphics[width=0.9\textwidth]{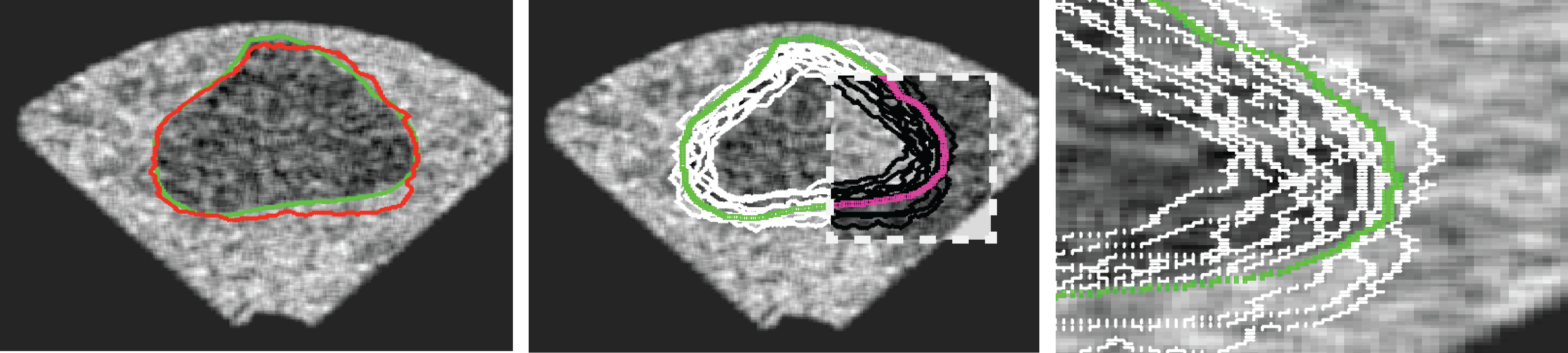}\vspace{0.07in}
\caption{Left: Sample image with gold segment and consensus contour; Middle: Simulated user  segments with the gold contour; Right: The inverted region (middle) magnified to show the variability.}
\label{fig:sampleImages}
\end{center}
\end{figure*}

We generated 500 images from their corresponding gold-standard images\footnote{All images and their segments are available online: http://tizhoosh.uwaterloo.ca/}. Furthermore, we generated 20 different segments for each image, assuming that there were 20 users. Figure  \ref{fig:TRUSsample} shows five examples of real and synthetic TRUS images. One should bear in mind that the purpose here was not to simulate the images realistically, but rather to have a base from which to generate variable segments from a perfect segment. Figure \ref{fig:sampleImages} shows an example of the gold segments and simulated user contours. The variability, coupled with the gold segment, is what is needed in our experiments.    

\subsection{Experiments}
We conducted several experiments to examine the effect of employing the new accuracy measures. In the first experiments we measured the accuracy of all 10,000 segments (500 images each segmented by 20 simulated users). The accuracy measurement encompassed the conventional Jaccard index $\bar{J}$, the Jaccard values for the three zones $\bar{J}_{Z_1}$, $\bar{J}_{Z_2}$ and $\bar{J}_{Z_3}$, as well as the two variations of total Jaccard accuracies for the entire segments $\bar{J}_1^*$	 and $\bar{J}_2^*$. These results are reported in Table \ref{tab:allresulst75}. It is apparent the extended Jaccard values are lower than the conventional ones: $\bar{J}>\bar{J}_1^*>\bar{J}_2^*$. The selection of the best segment may change depending on the measure whereas zonal accuracies show a more pronounced shift. In particular, if one chooses $\bar{J}_{Z_2}$ (zone 2) as a base, the results may have a different impact with resect to the quality of the segments. Similar results were observed for Dice coefficient $D$ and its anatomy-aware version $D^*$.  
 
\begin{table}[]
\centering
\caption{All results for images with $J\!>\!75\%$. Highest accuracies are highlighted for each measure.}
\label{tab:allresulst75}
\begin{tabular}{c|c||ccc||ll}
User 	& $\bar{J}$	& $\bar{J}_{Z_1}$	& $\bar{J}_{Z_2}$	& $\bar{J}_{Z_3}$	& $\bar{J}_1^*$	& $\bar{J}_2^*$ \\ \hline
1    & 87$\pm$ 6 & 64$\pm$ 20 & \CL 83$\pm$ 12 & 61$\pm$ 22 & 82$\pm$ 10 & \CL 74$\pm$ 11 \\
2    & 82$\pm$ 5 & 48$\pm$ 17 & 78$\pm$ 13 & 50$\pm$ 19 & 74$\pm$ 9  & 67 $\pm$ 7  \\
3    & 83$\pm$ 4 & 63$\pm$ 13 & 75$\pm$ 11 & 53$\pm$ 19 & 77$\pm$ 7  & 67$\pm$ 8  \\
4    & 85$\pm$ 5 & 64$\pm$ 13 & 76 $\pm$ 11 & 57$\pm$ 20 & 79$\pm$ 7  & 70$\pm$ 9  \\
5    & 80$\pm$ 4 & 45$\pm$ 13 & 75$\pm$ 10 & 43$\pm$ 15 & 71$\pm$ 6  & 65$\pm$ 6  \\
6    & 81$\pm$ 4 & 59$\pm$ 14 & 72$\pm$ 10 & 49$\pm$ 16 & 74$\pm$ 6  & 64$\pm$ 7  \\
7    & 80$\pm$ 4 & 45$\pm$ 14 & 74$\pm$ 12 & 44$\pm$ 16 & 71$\pm$ 7  & 65$\pm$ 5  \\
8    & 86$\pm$ 7 & 54$\pm$ 23 & 82$\pm$ 14 & \CL 64$\pm$ 22 & 80$\pm$ 11 & 73$\pm$ 10 \\
9    & \CL 88$\pm$ 5 & \CL 66$\pm$ 17 & 81$\pm$ 11 & 63$\pm$ 21 & \CL 83$\pm$ 8  & \CL 74$\pm$ 9  \\
10   & 81$\pm$ 4 & 59$\pm$ 13 & 72$\pm$ 11 & 49$\pm$ 18 & 74$\pm$ 7  & 64$\pm$ 8  \\
11   & 87$\pm$ 6 & 64$\pm$ 20 & \CL 83$\pm$ 12 & 61$\pm$ 22 & 82$\pm$ 10 & \CL 74$\pm$ 11 \\
12   & 82$\pm$ 5 & 48$\pm$ 17 & 78$\pm$ 13 & 50$\pm$ 19 & 74$\pm$ 9  & 67$\pm$ 7  \\
13   & 83$\pm$ 4 & 62$\pm$ 13 & 75$\pm$ 11 & 53$\pm$ 19 & 77$\pm$ 7  & 67$\pm$ 8  \\
14   & 85$\pm$ 5 & 64$\pm$ 13 & 76$\pm$ 11 & 57$\pm$ 20 & 79$\pm$ 7  & 70$\pm$ 9  \\
15   & 80$\pm$ 4 & 45$\pm$ 13 & 75$\pm$ 10 & 43$\pm$ 15 & 71$\pm$ 6  & 65$\pm$ 5  \\
16   & 81$\pm$ 4 & 59$\pm$ 14 & 72$\pm$ 10 & 49$\pm$ 16 & 74$\pm$ 6  & 64 $\pm$ 7  \\
17   & 80$\pm$ 4 & 45$\pm$ 15 & 75$\pm$ 11 & 44$\pm$ 16 & 71$\pm$ 7  & 65$\pm$ 5  \\
18   & 86$\pm$ 7 & 54$\pm$ 23 & 82$\pm$ 14 & \CL 64$\pm$ 22 & 80$\pm$ 11 & 73 $\pm$ 10 \\
19   & \CL 88$\pm$ 5 & \CL 66$\pm$ 17 & 81$\pm$ 11 & 63$\pm$ 21 & \CL 83$\pm$ 8  & \CL 74$\pm$ 9  \\
20   & 81$\pm$ 4 & 59$\pm$ 13 & 72$\pm$ 11 & 49$\pm$ 18 & 74$\pm$ 7  & 64 $\pm$ 8 
\end{tabular}
\end{table}

As a subset of the experiments, we randomly selected 50 images and 10 simulated users to examine some details (see Table \ref{tab:allACC}).   Both versions of anatomy-aware Jaccard deliver lower accuracies for any given user. Whereas $J_1^*$ is on average 10\% lower, $J_2^*$ is about 16\% lower.  The zone 1 seems to be the most difficult zone for almost all users. However, some users (e.g., users 1, 3, 4 and 6) appear to be more challenged with the zone 3.  Users 8 and 9 are the best users ($\bar{J}=$ 87 and 86, respectively). Their performance, however, is quite low when segmenting the zone 1 ($\bar{J}_{Z_1}=$ 59 and 66, respectively). Their performance seems to be more plausibly captured by the first anatomy-aware measure ($\bar{J}_1^*=$ 81 and 82, respectively) which also favors user 9 instead of user 8.  The second anatomy-aware measure appears to be very conservative ($\bar{J}_2^*=$ 74 and 72, respectively). Both standard deviation and variance illustrate that user variability is amplified by variability in zones 1 and 3. The first anatomy-aware measure, $\bar{J}_1^*$, seems to more pronouncedly quantify the user variability. Table \ref{tab:rankings} shows how the ranking of users  change when we base our evaluations upon anatomy-aware measures. Apparently, the ranking of users with excellent segmentation skills may not change much. In contrast, considerable shift in ranking can be observed when the user skills is rather  average.  For users with high Jaccard value, the ranking does not seem to change (users 3, 4, 8 and 9). Users with poor segmentation skills (user 10) does not seem either to change their ranking. For users with ``average" skills (Jaccard values around 60\%-70\%), the ranking may considerably change if we use anatomy-aware Jaccard (gray rows in Table \ref{tab:rankings}).

\begin{table}[htb]
\caption{Accuracy measurements via conventional Jaccard (first column), the defined three zones (gray columns), and the two anatomy-aware versions of Jaccard (last two columns). }
\begin{center}
\begin{tabular}{|c|c||c|c|c||c|c|}
User	& $\bar{J}$	& $\bar{J}_{Z_1}$	& $\bar{J}_{Z_2}$	& $\bar{J}_{Z_3}$	& $\bar{J}_1^*$	& $\bar{J}_2^*$ \\
\hline
1	& 69	& \CL 46	& \CL 57	& \CL 44	& 60	& 51\\
2	& 74	& \CL 35	& \CL 68	& \CL 48	& 61	& 57 \\
3	& 79	& \CL 59	& \CL 66	& \CL 56	& 72	& 62 \\
4	& 78	& \CL 57	& \CL 65	& \CL 56	& 71	& 61 \\
5	& 70	& \CL 30	& \CL 59	& \CL 42	& 57	& 53 \\
6	& 72	& \CL 50	& \CL 59	& \CL 48	& 64	& 54 \\
7	& 72	& \CL 37	& \CL 66	& \CL 39	& 59	& 55 \\
8	& 87	& \CL 59	& \CL 80	& \CL 74	& 81	& 74 \\
9	& 86	& \CL 66	& \CL 76	& \CL 71	& 82	& 72 \\
10	& 57	& \CL 12	& \CL 45	& \CL 18	& 36	& 39 \\ \hline
STDV	& 9	& 17	& 10	& 16	& 13	& 10 \\ \hline
variance	& 69	& 248	& 88	& 234	& 161	& 94 \\
\hline
\end{tabular}
\end{center}
\label{tab:allACC}
\end{table}%

\begin{table}[htb]
\caption{Ranking of segmentation skills of simulated users based on different accuracy measures.}
\begin{center}
\begin{tabular}{|c|c||c|c|c||c|c|}
Rank 	& $\bar{J}$	& $\bar{J}_{Z_1}$	& $\bar{J}_{Z_2}$	& $\bar{J}_{Z_3}$	& $\bar{J}_1^*$	& $\bar{J}_2^*$ \\
\hline
1	& 8	& 9	& 8	& 8	& 9	& 8 \\
2	& 9	& 8	& 9	& 9	& 8	& 9 \\
3	& 3	& 3	& 2	& 3	& 3	& 3 \\
4	& 4	& 4	& 3	& 4	& 4	& 4 \\
5	& \CL 2	& \CL 6	& \CL 7	& \CL 2	& \CL 6	& \CL 2 \\
6	& \CL 6	& \CL 1	& \CL 4	& \CL 6	& \CL 2	& \CL 7 \\
7	& \CL 7	& \CL 7	& \CL 6	& \CL 1	& \CL 1	& \CL 6 \\
8	& \CL 5	& \CL 2	& \CL 5	& \CL 5	& \CL 7	& \CL 5 \\
9	& \CL 1	& \CL 5	& \CL 1	& \CL 7	& \CL 5	& \CL 1 \\
10	& 10	& 10	& 10	& 10	& 10	& 10 \\ \hline
\end{tabular}
\end{center}
\label{tab:rankings}
\end{table}%

\section{Conclusions}
We introduced the novel idea of anatomy-aware accuracy measures. Extending commonly used measures such Jaccard index and Dice coefficient to anatomy-sensitive schemes is proposed by designing multiple necessary algorithms. Among others, the concept of ``master gold'' is introduced which is necessary for implementation of any anatomy-aware accuracy measurement.  
Anatomy-sensitive accuracy measurement appears to provide more insight into the challenges of medical image segmentation. By considering anatomical zones within segments, we may be able to develop a better understanding of contouring skills of users. As well, anatomy-aware accuracy measures seem to provide a more realistic qualification of inter-observe variability. And finally, anatomy-aware measures can be used to improve the performance of trainable segmentation accuracy \cite{Othman2014,Othman2013,Sahba2007}. 

\section*{Contributions}
The extensions of Jaccard and Dice measures to their zonal versions have been designed by the first author. As well, the image simulation to generate test data was designed and implemented by the first author. The second author has conducted all experiments and generated all results. The paper has been written by the first author. 

\section*{Acknowledgements}       
The authors would like to thank Dr. Masoom Haider (Sunnybrook Research Institute, Toronto) for some early discussions and advice with respect to the anatomy of the prostate gland. Also Dr. Haider provided us with some insight into the nature of the accuracy problem for the prostate gland. As well, the authors would like to thank Dr. Farzad Khalvati (Dept. of Medical Imaging, University of Toronto) for some initial elaborations on how experiments should be conducted. 

This project was funded by the Natural Sciences and Engineering Research Council of Canada (NSERC)  in form of a Discovery Grant.
 
\bibliography{ref}
\bibliographystyle{spiebib}

\newpage
\section*{Appendix}

\begin{algorithm}[h!tb]
\caption{Determine salient contour points on $M_G$ outline for mapping}
\label{alg:MasterGoldPoints}
\begin{algorithmic}[1]
\STATE Load the master gold $M_G$.
\STATE  Load the points $P_1,P_2,\dots,P_{N_Z}$.
\STATE  Copy the border points $P$ into $P_B$.
\STATE  Get the number of border points $N_{P_B}$.
\STATE \% --- Calculate relative border distances ---
\STATE Determine a starting point $C$ on $M_G$'s contour (see Figure \ref{fig:PointMapping}).
\STATE Get $M_G$'s contour, $(X_C, Y_C)$, starting from $C$.
\STATE Get the segment length $L = \max(X_C) - \min(X_C)$
\STATE Get the segment width $W = \max(Y_C) - \min(Y_C)$
\FOR{i = 1 : $N_{P_B}$}
	\STATE  Calculate the distance $D(i, 1)$ from $C$ to $P_B(x_i,y_i)$.
	\STATE  Normalize the distance $D(i, 2) = D(i, 1) / |X_C|$.
\ENDFOR
\STATE \% --- Calculate relative internal distances  ---
\STATE Copy the middle points from $P$ to $P_M$.
\FOR{i = 1 : $N_Z$}
	\STATE Determine a starting point $C_Z$ on the border of the $i$-th zone between the zone end points.
	\STATE Calculate the distance $D_W(i, 1) = ||C_{Z_i} , P_{M_i}||$.
	\STATE \% --- Normalize the distance --- 
	\IF{the zone on the right or on the left} 
		\STATE $S = W$
	\ELSE 
		\STATE $S = L$
	\ENDIF
	\STATE $DW(i, 2) = DW(i, 1)/S$.
\ENDFOR
\STATE Add $C_Z$ to $P$.
\STATE  Save $D,D_W,P$.
\end{algorithmic}
\end{algorithm}

\begin{algorithm}[h!tb]
\caption{Map zones to the ground-truth $G$}
\label{alg:Map2Gold}
\begin{algorithmic}[1]
\STATE Load $P,D,D_W$
\STATE Read the current ground-truth image $G$.
\STATE Determine a starting point $C_G$ on $G$'s contour (see Figure \ref{fig:PointMapping}).
\STATE Get $G$'s contour, $(X_G, Y_G)$, starting from $C_G$.
\STATE get the length $L_G$ and the width $W_G$ of $G$.
\STATE Calculate the distance $D_G$ from $C_{G_Z}$ to the suggested zone border points on $G$: $D_G = D(:, 2) \times length(X_G)$.
\STATE Calculate the point at the border of each zone $PGB$ on $G$ using $D_G$ and $(X_G, Y_G)$: $P_{G_B} = [X_G(D_G) Y_G(D_G)]$.
\STATE Calculate the centre points at the border of each zone $C_{G_Z}$ using $P_{G_B}$.
\STATE Calculate $S_G$ the same way as $S$.
\STATE Calculate the distance $D_{G_W}$ from $C_{G_Z}$ to the middle point of the zone: $D_{G_W} = D_W(:, 2)\times S_G$
\STATE Calculate the point at the curve of each zone $P_{G_M}$ using $C_{G_Z}$ and $D_{G_W}(:, 2)$.
\STATE Using $P_{G_B}$ and $P_{G_M}$, draw the curve of the zone.
\STATE Save the coordinates of the zone $(x_{G_i}, y_{G_i})$.
\STATE Save the polynomial parameters $P_N$ used for drawing the curve.
\end{algorithmic}
\end{algorithm}

\end{document}